# Calibration and transfer in indicator-based assessments of artificial consciousness

**Florentin Koch**

École Polytechnique, Institut Polytechnique de Paris, 91120 Palaiseau, France

Email: florentin.koch@polytechnique.edu

Article type: Spotlight Commentary

Author manuscript. Published as: Florentin Koch (2026), *Calibration and transfer in indicator-based assessments of artificial consciousness*. *Neuroscience of Consciousness*, 2026(1), niag061. Version of record: https://doi.org/10.1093/nc/niag061. This author version precedes the publisher's copyediting and typesetting.

## Abstract

Research on artificial consciousness increasingly shifts evaluation from behaviour to internal architecture. Theory-based indicators are used to update probability assignments. This improves on behavioural tests but raises two distinct problems. First, these assignments cannot currently be calibrated against independently established artificial consciousness outcomes. Second, their evidential relevance is transferred from biological cases without independent support that indicator-consciousness relations remain stable across substrates. This commentary distinguishes calibration from transfer and adapts the iterative natural-kind strategy by proposing a preliminary, theory-relative comparative space for cross-substrate assessment.

## Commentary

Butlin et al. (2025) provide an explicit formulation of the dominant approach to artificial consciousness. Because behavioural criteria can be imitated without guaranteeing subjective experience, evaluation shifts towards internal architecture. The authors derive indicators from recurrent processing, global workspace, higher-order and attention schema theories, supplemented by predictive processing, agency and embodiment. Under a computational-functionalist working assumption, each indicator alters the probability assigned to a system being conscious, while the combined profile informs the assessment. This advances beyond naive behaviourism.

The first problem is calibration in the artificial domain. Let $C_A$ denote that an artificial system is conscious and $I_A$ that it displays an indicator. The posterior $P(C_A \mid I_A)$ is the probability assigned to consciousness after observing it. In odds form, the posterior odds equal the prior odds multiplied by the indicator's likelihood ratio:

$$P(C_A \mid I_A) / P(\neg C_A \mid I_A) = P(C_A) / P(\neg C_A) \times P(I_A \mid C_A) / P(I_A \mid \neg C_A)$$

These probabilities represent epistemic degrees of belief, not assumed objective chances. Bayesian coherence concerns their internal consistency; predictive performance concerns how well they anticipate outcomes; calibration asks whether, among cases assigned probability x, approximately a proportion x realises the outcome. Assessing calibration therefore requires paired information about indicator presence and consciousness status in a defined reference class to constrain $P(C_A, I_A)$. Equivalently, this would constrain $P(I_A \mid C_A)$, $P(I_A \mid \neg C_A)$, and $P(C_A)$. Combining indicators also requires estimates of conditional dependencies and justified treatment of their source theories. Yet no artificial population currently provides independently established consciousness labels sufficient to identify these relations. The assignments may be rational, but their numerical values cannot be evaluated for calibration. Michel (2023) argues that a perfect phenomenal gold standard is not always required when partially trusted methods and contrasts can be compared. The artificial domain, however, offers no independently established conscious cases or generally accepted contrast class for testing them.

The Digital Consciousness Model (Shiller et al. 2026) illustrates this limitation. It combines expert assessments of indicator presence with priors and evidential relations drawn from theories and modelling choices. Its outputs may be coherent given those inputs, but cannot be assessed against independently established artificial consciousness outcomes. They formalise structured uncertainty without converting it into empirically calibrated probabilities. Rational probabilistic assessment remains possible, and uncertainty is not itself a defect: a forecast may be uncertain yet well calibrated. Here, the outcomes required to assess calibration are unavailable.

Calibration and transfer are distinct but connected. Direct calibration would require cases in which $C_A$ and $I_A$ can be jointly assessed. Because no artificial population provides independently anchored consciousness outcomes, indicator programmes cannot estimate the evidential force of $I_A$ within the artificial domain. They instead derive indicators from theories developed and tested primarily in biological systems, then ask whether artificial systems instantiate analogous properties. This responds to the absence of artificial consciousness labels, but shifts the epistemic burden: the assessment now depends on whether an indicator retains its evidential relation to consciousness beyond the biological domain.

The biological source relation is itself only partially constrained. Competing theories select different markers, including local recurrence, frontoparietal signatures and complexity measures, and assign them different roles. Even in humans, it is often unclear whether a marker tracks a neural correlate of consciousness proper, a prerequisite or a consequence of conscious processing, and whether it is necessary, sufficient or merely probabilistically associated with conscious states (Aru et al. 2012; Koch et al. 2016; Cleeremans, Mudrik and Seth 2025).

Even if this limitation were removed, transfer would remain problematic. Let $C_B$ and $I_B$ denote consciousness and indicator presence in biological systems. Suppose theories converged and their indicators were robustly validated across accessible biological cases, including infants, non-human animals and patients with disorders of consciousness. Biological evidence could then strongly constrain

$$P_B(C_B, I_B),$$

and the evidential relation between consciousness and indicator presence in biological systems. It would not establish that the relevant dependence structure is preserved in

$$P_A(C_A, I_A).$$

We would possess a well-supported biological relation, not evidence that artificial analogues preserve its relation to experience. Functional similarity may establish correspondence between biological and artificial properties, but not invariance in their evidential relation to consciousness. Applying a biological indicator to an artificial system therefore requires a bridge assumption specifying which features of that relation should carry over.

Computational functionalism can license transfer if the consciousness-relevant functional organisation is preserved. The remaining question is whether indicators establish that this organisation has been preserved, rather than only an operational analogue at an insufficient level of description. Biological-to-artificial induction is not impossible, and substrate difference does not defeat it by itself. Its evidential force remains conditional on a bridge principle that biological evidence does not independently validate.

This difficulty motivates a comparative strategy, not a defence of biological naturalism. Bayne et al.'s (2024) iterative natural-kind (NK) strategy validates consciousness tests by moving from consensus cases to neighbouring and progressively more alien populations, while revising confidence through convergence and divergence among tests. It leaves open how proximity between populations should be measured.

For cross-substrate transfer, this commentary offers a preliminary and revisable specification of that open question. Rather than placing systems on a single biological-to-digital continuum, it locates them in a theory-relative comparative space provisionally structured by physical realisation, causal organisation and dynamic regime. This is neither an alternative framework for consciousness-test validation nor a claim that these dimensions are uniquely correct, but an initial adaptation of the NK strategy intended to make transfer assumptions partially testable.

These dimensions are neither exhaustive nor independent, may be revised or supplemented, and depend on the theory or indicator examined. Physical realisation concerns the biological, electronic or hybrid mechanisms implementing computation. Causal organisation concerns connectivity and cross-scale relations. Dynamic regime concerns temporal, continuous-discrete and adaptive properties. Other dimensions, such as embodiment or self-regulation, may be added. Organoids, biohybrid systems, neuromorphic architectures and connectome-constrained simulations can therefore occupy different positions in this space rather than successive stages of a predetermined series (Figure 1).

Possible differences between biological and conventional digital computation motivate these comparisons. Milinkovic and Aru (2026) emphasise continuous-discrete dendritic integration, cross-scale interactions and metabolic constraints. Current digital architectures may not reproduce this combination. If some properties matter to consciousness, an abstractly similar function may fail to preserve what makes an indicator evidentially relevant. Kagan et al. (2022), who coupled in vitro cortical neurons to an artificial environment, and Dorkenwald et al. (2024), whose connectome supports biologically constrained modelling, show that experimentally tractable comparative systems can be constructed.

These studies do not establish consciousness, justify ordering systems by presumed proximity to it, or provide phenomenal ground truth. They show that selected features can be preserved or varied under controlled conditions to test specific transfer hypotheses. Comparative validation can examine whether indicator profiles remain convergent and aligned with established biological contrasts, and whether responses to theory-relevant interventions remain stable as physical realisation, causal organisation or dynamic regime varies.

Failures of invariance would weaken the corresponding transfer claim. Stability would satisfy a necessary robustness condition for that claim, without establishing preservation of the indicator-consciousness relation. It would not show that the indicators track consciousness, calibrate $P(C_A \mid I_A)$, or supply phenomenal ground truth. Comparative validation instead constrains the assumption that licenses treating an artificial indicator as evidence for artificial consciousness. It tests the robustness of the evidential relations on which that inference depends, decomposing the bridge assumption into partially testable comparisons without eliminating it.

Indicator-based programmes remain useful for organising heterogeneous evidence and moving beyond behavioural tests. Their numerical outputs should nevertheless be interpreted as assumption-sensitive probability assignments, not empirically calibrated probabilities of artificial consciousness. The priority is not to deny that current systems may be conscious, but to make explicit and test the bridge assumptions under which biological indicator-consciousness relations are transferred to artificial systems.

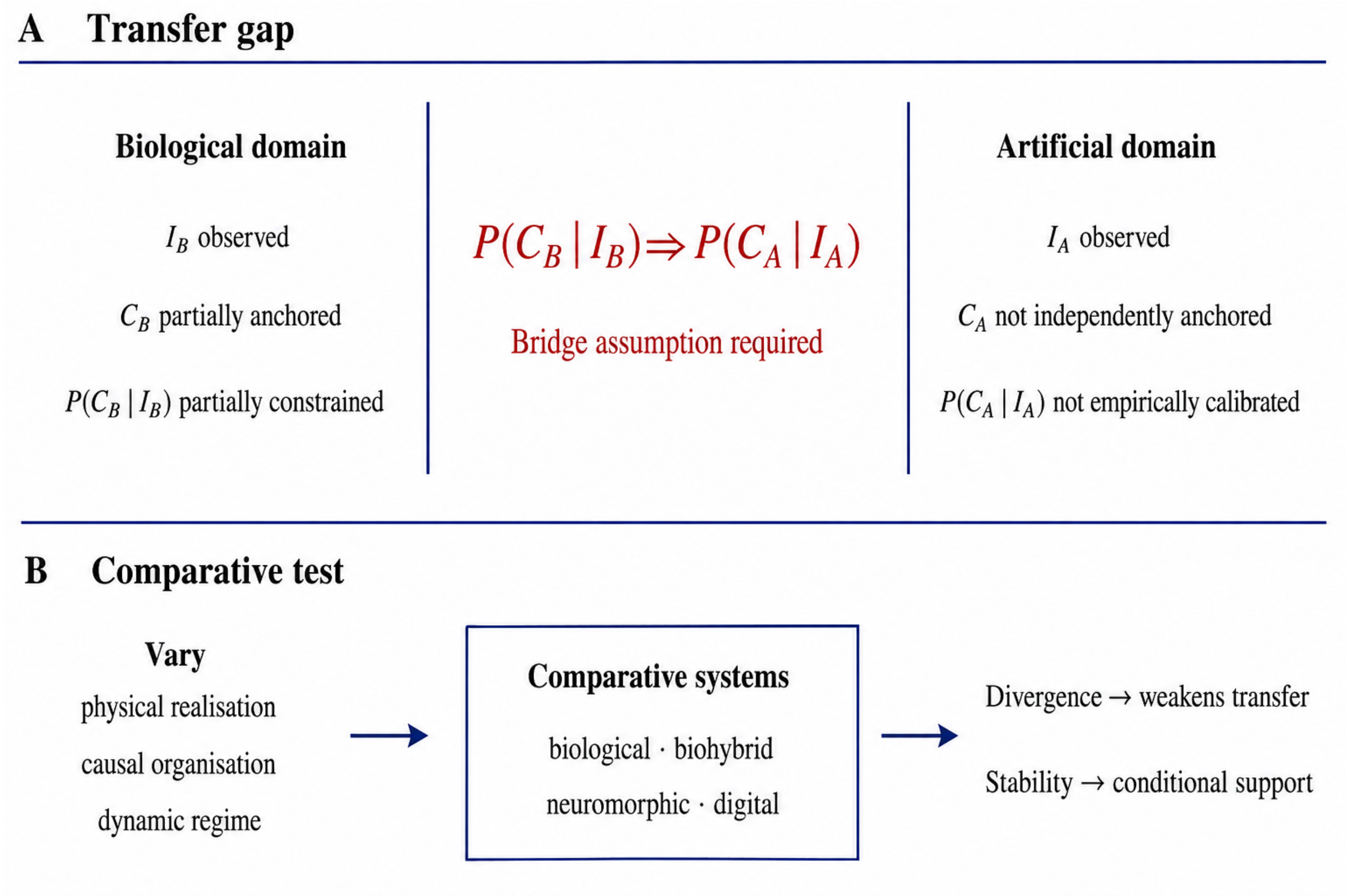

**Figure 1. Transfer** gap and comparative testing in indicator-based assessments of artificial consciousness. (A) In biological systems, indicator-consciousness relations are partially constrained because consciousness is at least partly anchored by converging evidence. In artificial systems, consciousness is not independently anchored, so the corresponding posterior cannot currently be empirically calibrated. **Transferring** the evidential relation from biology to artificial systems therefore requires an additional bridge assumption. (B) Comparative tests vary physical realisation, causal organisation and dynamic regime across biological, biohybrid, neuromorphic and digital systems. Divergence weakens the relevant **transfer** claim, whereas stability satisfies a robustness condition without calibrating artificial consciousness or supplying phenomenal ground truth.

***Alt text**: Schematic contrasting the calibration gap in artificial systems with comparative cross-substrate testing. Biological indicator-consciousness relations are partially constrained, whereas artificial assessments require a bridge assumption. Comparative systems vary in physical realisation, causal organisation and dynamic regime to test the robustness of this transfer.*

## Acknowledgements

OpenAI's ChatGPT was used for language editing and assistance with figure preparation. The author reviewed and revised all outputs and takes full responsibility for the final manuscript.

## Funding

The author received no specific funding for this work.

## Conflict of interest

None declared.

## Data availability

No new data were generated or analysed in support of this research.